\def\confName{CVPR}
\def\confYear{2027}

\def\paperTitle{DrawScore: Evaluating Generated CAD Models against Engineering Drawings}

\def\authorBlock{
    Haorui Chen$^{1}$\thanks{Equal contribution.} \quad
    Shiteng Lu$^{1}$\footnotemark[1] \quad
    Zhanyu Guo$^{1}$ \quad
    Zaixi Li$^{1}$ \quad
    Yunhe Ni$^{1}$ \\
    Yaosong Du$^{1}$ \quad
    Yichen Xiang$^{1}$ \quad
    Xiande Huang$^{2}$ \quad
    Rui Huang$^{3}$\thanks{Corresponding author.} \\[0.5em]
    {\normalsize $^{1}$University of Electronic Science and Technology of China} \\
    {\normalsize $^{2}$De Artificial Intelligence Technology Co., Ltd} \\
    {\normalsize $^{3}$University of Hong Kong} \\[0.3em]
    {\tt\small \{hrchen,lushiteng,huang\_rui\}@std.uestc.edu.cn}
}

\newif\ifreview 
\newif\ifarxiv \newcommand{\arxiv}{\arxivtrue}
\newif\ifcamera 
\newif\ifrebuttal 

\arxiv 

\documentclass[10pt,twocolumn,letterpaper]{article}
\PassOptionsToPackage{table}{xcolor}
\ifreview \usepackage[review]{cvpr} \fi
\ifarxiv \usepackage[pagenumbers]{cvpr} \fi
\ifrebuttal \usepackage[rebuttal]{cvpr} \fi
\ifcamera \usepackage{cvpr} \fi

\usepackage[T1]{fontenc}
\usepackage{graphicx}	
\usepackage{amsmath}	
\usepackage{amssymb}	
\usepackage{booktabs}
\usepackage{times}
\usepackage{microtype}
\usepackage{epsfig}
\usepackage{caption}
\usepackage{float}
\usepackage{placeins}
\usepackage{color, colortbl}
\usepackage{stfloats}
\usepackage{enumitem}
\usepackage{tabularx}
\usepackage{xstring}
\usepackage{multirow}
\usepackage{xspace}
\usepackage{url}
\usepackage{subcaption}
\usepackage[hang,flushmargin]{footmisc}

\usepackage{textcomp}
\usepackage{adjustbox}
\usepackage{listings}
\usepackage{algorithm}
\usepackage{algpseudocode}
\usepackage{pifont}
\usepackage[most]{tcolorbox}
\definecolor{findingback}{RGB}{247,250,255}
\definecolor{findingframe}{RGB}{96,132,180}
\newtcolorbox{findingbox}[1][]{%
    enhanced,
    colback=findingback,
    colframe=findingframe!70,
    boxrule=0.35pt,
    arc=3pt,
    left=5pt,right=5pt,top=3pt,bottom=3pt,
    before skip=4pt,after skip=4pt,
    #1
}

\newcommand{\bi}[1]{\textbf{\textit{#1}}}

\ifcamera \usepackage[accsupp]{axessibility} \fi

% ...

\ifarxiv  \fi

\newcommand{\R}[1]{{%
    \textbf{%
        \ifstrequal{#1}{1}{\textcolor{red}{R#1}}{%
        \ifstrequal{#1}{2}{\textcolor{blue}{R#1}}{%
        \ifstrequal{#1}{3}{\textcolor{magenta}{R#1}}{%
        \ifstrequal{#1}{4}{\textcolor{teal}{R#1}}{%
                           \textcolor{cyan}{R#1}%
        }}}}%
    }%
}}

\usepackage{xr-hyper}

\makeatletter
\newcommand*{\addFileDependency}[1]{
  \typeout{(#1)}
  \@addtofilelist{#1}
  \IfFileExists{#1}{}{\typeout{No file #1.}}
}

\makeatother
\newcommand*{\myexternaldocument}[1]{
    \externaldocument{#1}
    \addFileDependency{#1.tex}
    \addFileDependency{#1.aux}
}

\definecolor{cvprblue}{rgb}{0.21,0.49,0.74}
\usepackage[pagebackref,breaklinks,colorlinks,allcolors=cvprblue]{hyperref}
\usepackage[capitalize]{cleveref}
\crefname{section}{Sec.}{Secs.}
\crefname{table}{Table}{Tables}
\crefname{figure}{Fig.}{Figs.}

\ifarxiv \crefname{appendix}{App.}{Apps.}
\else \crefname{appendix}{Suppl.}{Suppls.} \fi

\unless\ifarxiv \myexternaldocument{_supplementary} \fi

\begin{document}
\title{\paperTitle}
\author{\authorBlock}
\maketitle

\begin{abstract}
Automatic CAD generation is commonly evaluated with volumetric IoU, Chamfer Distance, and surface F-score, which barely penalize local errors that occupy only a small fraction of a part's volume or surface area, such as a misplaced hole or a slot on the wrong face, even though such errors are critical to manufacturing. We formalize these metrics in a unified integral form and show that the weight of a local error decays as the $(k+q)$-th power of its size, where $k$ is the dimension of the integration domain and $q$ is determined by the penalty function. Guided by this analysis, we propose DrawScore, which projects both models into three orthographic views with visible and hidden lines, as in engineering drawings, and applies a saturating penalty along the projected curves, so that the weight of a local error decays only linearly. Controlled experiments confirm the predicted decay orders, and on outputs of existing CAD generation agents, DrawScore exposes errors in outputs that IoU rates as nearly perfect. DrawScore further improves generation as a reinforcement learning reward and, for CAD generation from engineering drawings, serves as a test-time verifier without 3D ground truth or an additional reward model. Our code is available at \url{https://github.com/UESTC-Simon/DrawScore}.
\end{abstract}
\section{Introduction}
\label{sec:introduction}

Computer-aided design (CAD) models are the starting point for manufacturing almost every part.
With the widespread adoption of large language models and coding agents, automatic CAD generation has become an active research direction~\cite{guan2025cadcoder,hu2026itercad}.
In CAD generation, a reliable evaluation metric can both identify errors and serve directly as a reward to improve generation~\cite{guan2025cadcoder}.\footnote{A detailed discussion of related work is provided in \Cref{sec:related}.}
However, commonly used CAD geometric similarity metrics, including volumetric IoU~\cite{mescheder2019occupancy}, Chamfer Distance~\cite{fan2017pointset}, surface F-score~\cite{tatarchenko2019singleview}, and Edge Chamfer Distance~\cite{chen2020bspnet}, measure differences over a part's volume, surface, or sharp edges.
These metrics respond strongly when a part's overall size or shape is wrong, but give little weight to a missing or misplaced small feature, such as a hole or a slot, even though such errors matter for manufacturing.

\begin{figure}[t]
\centering
\includegraphics[width=\linewidth]{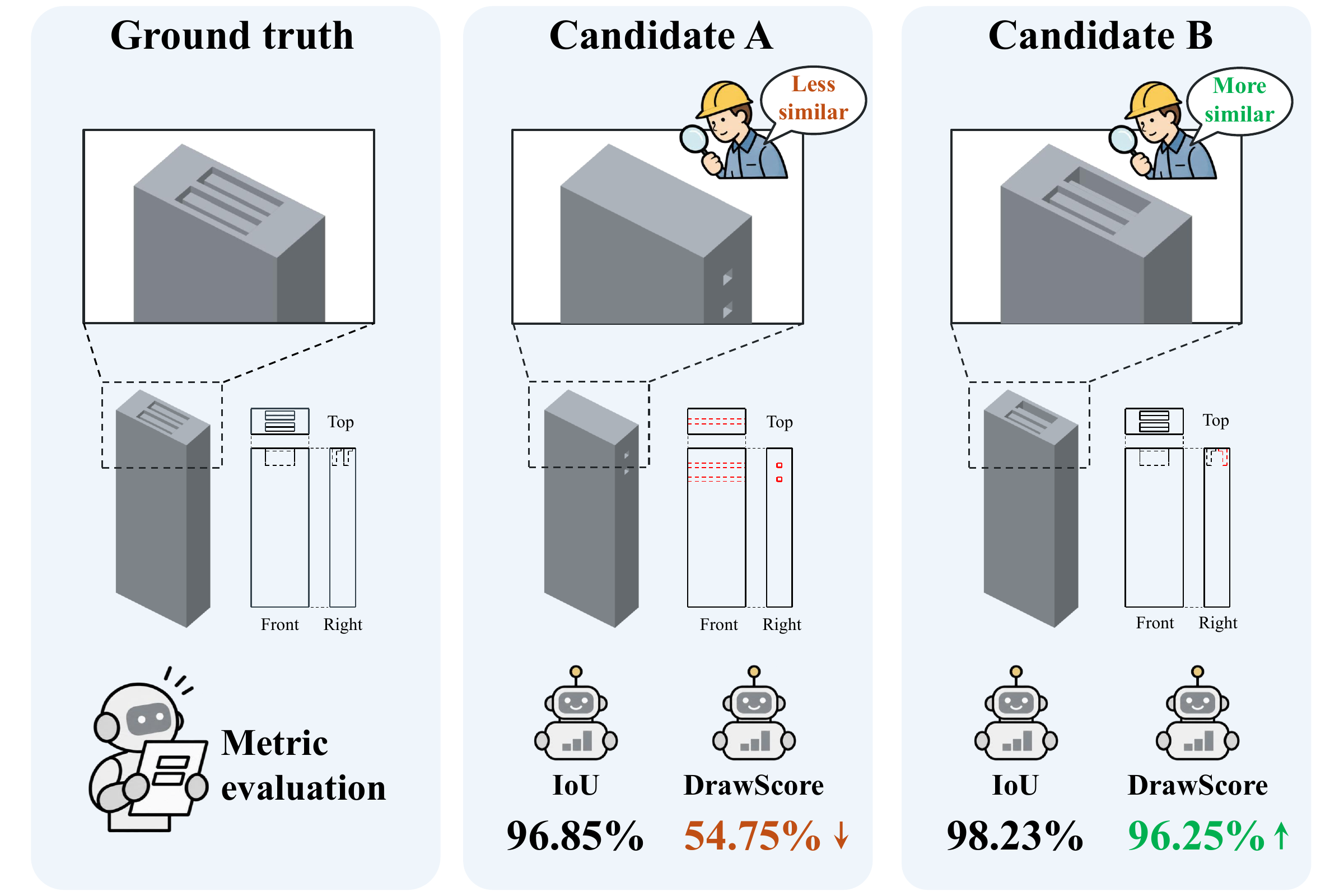}
\caption{Comparison of two candidate CAD models generated by an existing CAD generation agent against the ground truth.}
\label{fig:motivation}
\vspace{-12pt}
\end{figure}

\Cref{fig:motivation} illustrates how different metrics capture different aspects of geometric error.
Candidate A moves the two slots from the inclined top face to the right side face, whereas candidate B only reverses the profile of one slot.
Although candidate B is clearly closer to the ground truth, the two candidates receive similar IoU scores.
This is because IoU measures how much two solids overlap in volume.
Since the slots occupy only a small fraction of the part's volume, placing them on the wrong face changes the overlapping volume only slightly.
Surface-based metrics such as Chamfer Distance capture another aspect, namely how far the surfaces deviate, and likewise give little weight to errors on small surface areas.
In engineering drawings, the two errors above appear very differently: the error of candidate A changes the slot outlines in two views, whereas that of candidate B is confined to a small local region.

This example motivates us to evaluate CAD models through engineering drawings.
\ding{182} \emph{Engineering drawings align with how engineers inspect CAD models.}
After completing a model, engineers typically project it back into engineering drawings and compare them with the design drawings to identify errors.
Three orthographic views with hidden lines are the standard way to convey the geometry of a mechanical part~\cite{solidworks2011teaching,hu2023plankassembly}, so comparing these views captures most geometric differences between two models.
\ding{183} \emph{Engineering drawings capture a different aspect of geometric error.}
A small feature such as a slot occupies little volume or surface area but still leaves complete lines in the drawings, where its errors stand out.
We explain this difference through an integration dimension analysis in \Cref{sec:analysis}: volumetric and surface metrics integrate over 3D solids and 2D surfaces, respectively, whereas comparison through engineering drawings integrates over 1D projected curves.
Unlike the sharp edges of Edge Chamfer Distance, our curves are projected from the entire model and thus include smooth contours.
\ding{184} \emph{Engineering drawings support test-time scaling (TTS).}
Drawings are the standard medium for communicating designs in manufacturing~\cite{hu2023plankassembly}. In CAD generation from drawing~\cite{hu2023plankassembly,drawing2cad,cad2program,creftcad,cmecad,ortho2cad,omnimech}, the input drawing remains available at inference time, allowing direct comparison with a candidate's projected views without 3D ground truth or an additional trained reward model; the metric can therefore serve as a test-time verifier.  

We therefore propose \bi{DrawScore}, which projects the candidate and the ground truth into three orthographic views with visible and hidden lines and measures how well their projected curves coincide.
By capturing an aspect of geometric error different from that of existing metrics, DrawScore complements them for a more comprehensive assessment of generated CAD models.
We evaluate DrawScore on controlled samples and outputs from existing CAD generation agents, and use it as a reward for training and a verifier for TTS.
The results show that:
\ding{182} \textit{The decay order of each metric agrees with our integration dimension analysis.}
Experiments on controlled samples confirm that each metric decays at the order determined by its integration domain and penalty function (\Cref{sec:analysis}).
DrawScore also detects errors that other curve-based metrics miss.
\ding{183} \textit{A perfect volume does not imply a correct model, and the remaining errors of existing agents concentrate on small features.}
Among 1,188 outputs from six agent configurations, 18.9\% of those with IoU $\geq 0.95$ have a DrawScore below 0.95, whereas only 1.2\% of those with DrawScore $\geq 0.95$ have an IoU below 0.95.
Agents thus mostly get the volume right, and the remaining errors lie on the lines of the drawings.
\ding{184} \textit{As a reward, DrawScore improves generation quality across metrics.}
It raises the share of valid outputs from 75.2\% to 89.5\% and improves every reported metric over the SFT model, whereas the Edge Chamfer Distance reward, which ignores smooth contours, lowers DrawScore.
\ding{185} \textit{DrawScore selects candidates at inference time without 3D ground truth.}
For CAD generation from engineering drawings, selecting the best of $N$ candidates by their DrawScore against the input drawing outperforms single generation on most metrics, so DrawScore can serve as an inference-time proxy for them.


\section{Integration Dimension Analysis of Geometric Similarity Metrics}
\label{sec:analysis}

\begin{table*}[t]
\centering
\caption{Common geometric similarity metrics as instances of \cref{eq:unified}.
\textit{Score}: aggregation of the two directions; lower is better for CD and ECD.
\textit{Decay order}: the weight of a local error of size $s$ scales as $(s/L)^{\text{order}}$;
$^*$Decreases to 1 if no corresponding curve lies near the error.
\textit{Supports TTS}: whether the metric can score candidates against an input drawing.}
\label{tab:metrics}
\footnotesize
\setlength{\tabcolsep}{3pt}
\renewcommand{\arraystretch}{1.15}
\begin{tabular*}{\textwidth}{@{\extracolsep{\fill}}l l c c c c@{}}
\toprule
Metric & Integration domain & Penalty $\rho(d)$ & Score & Decay order & Supports TTS \\
\midrule
Volumetric IoU~\cite{mescheder2019occupancy} & Solid ($k{=}3$) & $\mathbf{1}[d>0]$ & $RP/(R+P-RP)$ & 3 & \ding{55} \\
Chamfer Distance~\cite{mescheder2019occupancy} & Surface ($k{=}2$) & $d/L$ & $D_{\rightarrow}+D_{\leftarrow}$ & 3 & \ding{55} \\
Surface F-score~\cite{tatarchenko2019singleview} & Surface ($k{=}2$) & $\mathbf{1}[d\geq\tau]$ & $2RP/(R+P)$ & 2 & \ding{55} \\
Edge Chamfer Distance~\cite{chen2020bspnet} & Model edges ($k{=}1$) & $d/L$ & $D_{\rightarrow}+D_{\leftarrow}$ & 2$^*$ & \ding{55} \\
\midrule
\textbf{DrawScore (ours)} & Projected curves ($k{=}1$) & $1-e^{-d^2/(2\sigma^2)}$ & $\min_v 2R_vP_v/(R_v+P_v)$ & 1 & \ding{51} \\
\bottomrule
\end{tabular*}
\vspace{-10pt}
\end{table*}

\subsection{Unified Integral Formulation}
\label{sec:analysis:formulation}

The metrics in \cref{tab:metrics} compare two models in the same way: each model is represented as a point set, the discrepancy of every point to the other model is measured, and the results are averaged.
The metrics differ in the dimension $k$ of this point set, which is a solid, a surface, or a set of curves.
Since a metric integrates over this $k$-dimensional point set, we call it the \emph{integration domain}.

Let $X$ and $Y$ denote the integration domains of the ground truth $G$ and the candidate $C$, and let $\mu$ denote the corresponding $k$-dimensional measure, \ie, volume, area, or length.
The directional discrepancy from $X$ to $Y$ is
\begin{equation}
  D(X \to Y) = \frac{1}{\mu(X)} \int_X \rho\bigl(d(x, Y)\bigr)\, d\mu(x),
  \label{eq:unified}
\end{equation}
where $d(x, Y) = \min_{y \in Y} \|x - y\|_2$ is the distance from $x$ to $Y$ and $\rho$ is a penalty function.
The penalty function maps this distance to a nonnegative penalty, which is 0 at zero distance and increases with distance.
$D(X \to Y)$ is thus the average penalty over $X$.

Swapping $X$ and $Y$ gives the opposite discrepancy $D(Y \to X)$.
We write $D_{\rightarrow} = D(X \to Y)$ and $D_{\leftarrow} = D(Y \to X)$, and $R = 1 - D_{\rightarrow}$ and $P = 1 - D_{\leftarrow}$, which correspond to recall and precision for saturating penalties.
An aggregation function $A$ combines the two directions into the final score.
Each metric is therefore determined by its integration domain, penalty function, and aggregation function, as listed in \cref{tab:metrics}.
Penalty functions fall into two types.
A \bi{growing} penalty is proportional to distance, such as $d/L$ in Chamfer Distance, where $L$ is the diagonal length of the ground truth's bounding box.
A \bi{saturating} penalty reaches its upper bound of 1 once the distance exceeds a tolerance and does not increase further, such as $\mathbf{1}[d \ge \tau]$ in surface F-score.
For simplicity, we analyze surface F-score in its distance-only form~\cite{tatarchenko2019singleview}.
For example, the penalty $\mathbf{1}[d > 0]$ of volumetric IoU is a saturating penalty with zero tolerance, so $R$ is the fraction of $G$'s volume inside $C$, and the score in \cref{tab:metrics} recovers $\mu(X \cap Y) / \mu(X \cup Y)$.

Combining the two directions does not change how the score scales with error size, so the following analysis considers a single direction and writes $D = D_{\rightarrow}$.

\subsection{Local Error Decomposition}
\label{sec:analysis:decomposition}

We call an error \emph{local} if it changes only part of a model's geometry while the rest coincides with the ground truth; its size $s$ is the scale of the changed region, such as the diameter of a hole.
Let $E_1, \dots, E_n \subset X$ be the disjoint regions of the integration domain changed by the errors, which we call \emph{error regions}.
Every point of $X$ outside the error regions has a coinciding point in the other model, \ie, $d(x, Y) = 0$, so its penalty is zero.
\Cref{eq:unified} therefore decomposes exactly into
\begin{equation}
  w_i = p_i\, \bar\rho_i, \qquad D = \sum_{i=1}^{n} w_i,
  \label{eq:decomposition}
\end{equation}
where $p_i = \mu(E_i) / \mu(X)$ is the \emph{error fraction} of error $i$, \ie, the volume, area, or length of $E_i$ as a share of the entire integration domain $X$, and $\bar\rho_i$ is the \emph{average penalty} over $E_i$.
The product $w_i$ is the \emph{weight} of error $i$ in $D$, \ie, how much the error lowers the score.

The error fraction depends only on which part of the integration domain an error changes, whereas the weight also depends on how heavily the changed points are penalized.
For saturating penalties, $\bar\rho_i \le 1$, and hence
\begin{equation*}
  w_i \le p_i, \qquad D \le \sum_{i=1}^{n} p_i,
\end{equation*}
\ie, the weight of an error never exceeds its error fraction.
Equality holds when the points in the error region deviate beyond the tolerance, since each of them then receives a penalty of 1.
For growing penalties, the average penalty is proportional to the deviation, so the weight is usually far smaller than the error fraction.

\begin{figure*}[t]
\centering
\IfFileExists{figs/sensitivity_image.pdf}{%
  \includegraphics[width=\textwidth]{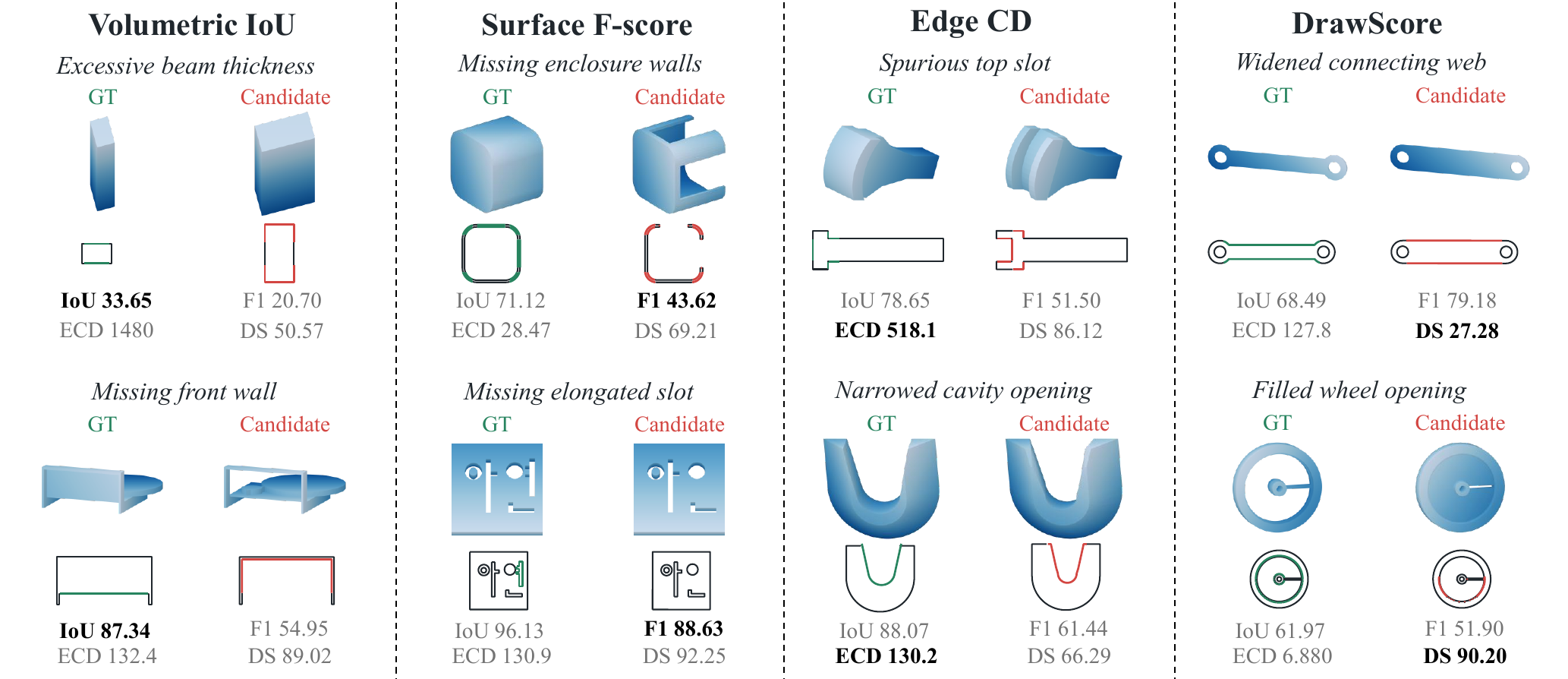}%
}{%
  \fbox{\parbox[c][35mm][c]{0.95\textwidth}{\centering
  Metric sensitivity figure placeholder.\\
  Add the final artwork as \texttt{figs/sensitivity_image.pdf}.}}%
}
\caption{Sensitivity of geometric similarity metrics to different local errors.
In each column, the top row shows an error to which the metric in \textbf{bold} is sensitive, and the bottom row an error to which it is insensitive.}
\label{fig:sensitivity}
\vspace{-10pt}
\end{figure*}

As shown in \cref{fig:sensitivity}, for DrawScore, widening the connecting web displaces its long outlines, which make up a large fraction of the projected curves, whereas filling the wheel opening changes a large volume but leaves the circles in the three views almost unchanged.
Thus no metric is uniformly more or less sensitive; each responds only to the fraction an error occupies in its own integration domain.

\Cref{eq:decomposition} shows that the weight $w_i$ is determined jointly by the error fraction $p_i$ and the average penalty $\bar\rho_i$.
The next two sections analyze these two factors in turn.

\subsection{The Integration Domain Determines the Error Fraction}
\label{sec:analysis:domain}

\paragraph{Dimension.}
Suppose local error $i$ is scaled down by a factor $t \in (0, 1]$ while the rest of the geometry stays fixed, and let $p_i(t)$ denote its error fraction.
The error region shrinks accordingly, so its measure scales as $t^k$, while $\mu(X)$ barely changes; hence
\begin{equation}
  p_i(t) \approx t^k\, p_i(1).
  \label{eq:share_scaling}
\end{equation}

\paragraph{Choice of curves.}
Solids and surfaces determine a model uniquely, so any geometric error changes them, and its error fraction $p_i$ is always non-zero.
An integration domain made of curves ($k=1$), in contrast, contains only some of the model's lines, and $p_i > 0$ only if the lines at the error are included.
For example, Edge Chamfer Distance takes only edges where faces meet at a distinct angle, so errors at smooth transitions such as fillets go undetected.

\begin{findingbox}
\textbf{Finding 1:} The error fraction of a local error scales as $t^k$, so it decreases more slowly in a lower-dimensional integration domain.
On curves, an error is detected only if the integration domain includes the lines at the error.
\end{findingbox}

\subsection{The Penalty Function Determines the Average Penalty}
\label{sec:analysis:penalty}

By \cref{eq:unified}, the penalty of a point $x$ in an error region is $\rho\bigl(d(x, Y)\bigr)$: the nearest distance $d$ is computed first and then converted into a penalty by $\rho$.
Therefore, when an error is scaled down by $t$, the change in its average penalty depends on two things: whether the distance shrinks with it, and how $\rho$ responds to the change in distance.

\paragraph{Distance.}
Whether the distance shrinks with the error depends on whether the other model has corresponding geometry near the error.
For the narrowed cavity opening in \cref{fig:sensitivity}, the candidate still has an opening, so the distance equals the width deviation and shrinks with the error.
For the missing elongated slot, no candidate line lies near the slot's lines, so the distance is set by the surrounding geometry and does not shrink.
Since solids and surfaces are continuous, they always pass near the error, so the latter case arises only in integration domains made of curves.

\paragraph{Penalty function.}
For a growing penalty $\rho(d) = (d/L)^q$, if the other model has corresponding geometry near the error, the distance shrinks by a factor of $t$ and the average penalty by $t^q$; otherwise, the distance stays unchanged and so does the average penalty.
For a saturating penalty, every point keeps a penalty of 1 as long as the distance exceeds the tolerance, so the average penalty stays unchanged.
Substituting the average penalty and the error fraction in \cref{eq:share_scaling} into \cref{eq:decomposition} gives how the weight of error $i$ varies with its size:
\begin{equation}
  w_i(t) \approx t^{k+q}\, w_i(1).
  \label{eq:scaling}
\end{equation}

\Cref{tab:metrics} lists the decay order of each metric given by \cref{eq:scaling}.
Although Chamfer Distance is defined on surfaces, its growing penalty makes its weight decrease as $t^3$, as fast as that of volumetric IoU.
Edge Chamfer Distance is the only metric that is defined on curves and uses a growing penalty, so its decay order depends on whether a corresponding line exists near the error: if so, its weight decreases as $t^2$; if not, only as $t$.
Even then, its penalty $d/L$ is usually far below 1, so its weight remains clearly lower than that of DrawScore.

\begin{findingbox}
\textbf{Finding 2:} The weight of a local error scales as $t^{k+q}$.
A growing penalty speeds up this decrease only when the other model has corresponding geometry near the error, which may not hold on curves; a saturating penalty never does.
\end{findingbox}

In summary, to preserve the weight of local errors, a metric should use an integration domain made of curves that includes the lines of the part as completely as possible, together with a saturating penalty.
DrawScore is designed according to these three principles.

\section{DrawScore}
\label{sec:method}

\subsection{Three Orthographic Views}
\label{sec:method:projection}

We use the CAD kernel's hidden-line removal algorithm to project the ground truth $G$ and the candidate $C$ along three orthogonal directions into views $v \in \{\text{front}, \text{top}, \text{right}\}$.
Each view contains the projections of the model's edges and the contours of smooth surfaces.
Let $G_v^{\text{vis}}$ and $G_v^{\text{hid}}$ denote the visible and hidden curves of the ground truth in view $v$, $C_v^{\text{vis}}$ and $C_v^{\text{hid}}$ their candidate counterparts, and $G_v = G_v^{\text{vis}} \cup G_v^{\text{hid}}$ and $C_v = C_v^{\text{vis}} \cup C_v^{\text{hid}}$ all curves in the view.

\subsection{Penalty Function}
\label{sec:method:penalty}

Visible and hidden lines are compared separately.
For a point $x$ on a ground-truth curve in view $v$, let $d(x, C_v)$ denote its distance to the candidate curves of the same visibility.
DrawScore uses a Gaussian saturating penalty:
\begin{equation}
  \rho(d) = 1 - \exp\!\left(-\frac{d^2}{2\sigma^2}\right),
  \label{eq:drawscore_penalty}
\end{equation}
where $\sigma$ is the positional tolerance.
The penalty is close to 0 when $d \ll \sigma$ and close to 1 when $d \gg \sigma$, no longer increasing with distance.

\subsection{Recall and Precision}
\label{sec:method:pr}

In each view, recall and precision (\cref{sec:analysis:formulation}) are computed with arc length as the measure:
\begin{equation}
  R_v = 1 - \frac{1}{\mathrm{len}(G_v)} \int_{G_v} \rho\bigl(d(x, C_v)\bigr)\, ds,
  \label{eq:recall}
\end{equation}
where $\mathrm{len}(\cdot)$ denotes total arc length, and $P_v$ is obtained by exchanging the ground truth and the candidate.

Before integration, we merge overlapping curves within each view.
Because integration follows arc length, the score does not depend on how curves are segmented, and hence not on how the model was constructed.

\subsection{Multi-View Aggregation}
\label{sec:method:aggregation}

As in surface F-score, the score of each view is the harmonic mean $S_v = 2P_vR_v/(P_v + R_v)$ of recall and precision, and the final score is the lowest score across the three views, $S = \min_v S_v$.

Some errors are visible only in specific views.
For example, if a round hole is modeled as a square hole of the same width, the hole appears as the same pair of hidden lines in the front and right views, and only the top view reveals the difference.
Averaging over the three views would let the two error-free views dilute this error to one third, whereas taking the minimum preserves the full difference in the top view.

\subsection{Weights of Local Errors}
\label{sec:method:analysis}

Writing $a_v = 1 - R_v$ and $b_v = 1 - P_v$ for the two directional discrepancies in view $v$, the definitions of $S_v$ and $S$ give
\begin{equation}
\begin{aligned}
  1 - S_v &= \frac{a_v + b_v - 2a_v b_v}{2 - a_v - b_v}
           \approx \frac{a_v + b_v}{2},\\
  1 - S   &= \max_v\, (1 - S_v).
\end{aligned}
\label{eq:drawscore_response}
\end{equation}
Thus, to first order, the score drop in each view is the average of the two directional discrepancies and likewise decomposes into the weights of local errors by \cref{eq:decomposition}, so the analysis of \cref{sec:analysis} applies directly to DrawScore.


\providecommand{\RQOneNumCases}{35}
\providecommand{\RQOneNumComparisons}{53}

\section{Experimental Setup}
\label{sec:experiments_setup}

\subsection{Research Questions}
\label{sec:experiments:rq_overall}

\textbf{\textit{RQ1: Does the integration dimension analysis hold?}}
We inject controlled local errors into ground-truth models and test three conclusions of \cref{sec:analysis} in turn.
We construct \RQOneNumComparisons{} comparisons from \RQOneNumCases{} such models.
The three conclusions are the decay order $k+q$ of \cref{eq:scaling}, the choice of curves in the integration domain, and the saturating penalty.
For the decay order, we scale one error over several sizes and measure how its weight varies with its size; the other two conclusions are each tested by representative errors that isolate one mechanism.

\textbf{\textit{RQ2: Do DrawScore and existing metrics capture different aspects of geometric error?}}
Taking IoU as an example, we select outputs that one metric scores nearly perfectly, plot the histogram of the other metric on them, and analyze the errors at different quantiles, in both directions.
We evaluate six configurations covering different models and agent frameworks, which yield 1{,}188 valid outputs.

\textbf{\textit{RQ3: Can DrawScore serve as a training reward that improves generation quality?}}
We use DrawScore as a reinforcement learning reward to train Qwen3.5-4B and compare it with rewards from other metrics defined on solids, surfaces, and curves.
The training and evaluation data follow Ortho2CAD~\cite{ortho2cad}.

\textbf{\textit{RQ4: Can DrawScore improve generation quality at inference time?}}
In CAD generation from engineering drawings, the input drawing is available at inference time.
We use DrawScore to compare multiple candidates with the input drawing and select the one with the highest score.

\subsection{Compared Metrics}
\label{sec:experiments:metrics}

We compute the five metrics in \cref{tab:metrics} as follows.
\ding{182}~\emph{Volumetric IoU (IoU)} is computed by Boolean operations on tessellated solids.
\ding{183}~\emph{Chamfer Distance (CD)} samples 50{,}000 points on the surface of each model, with distances normalized by $L$; we report it multiplied by $10^3$.
\ding{184}~\emph{Surface F-score (F1)} uses the same surface samples with the threshold $\tau = 0.005L$.
To match the analysis of \cref{sec:analysis}, RQ1 uses the distance-only form in \cref{tab:metrics}.
In RQ2--RQ4, following CADGenBench~\cite{cadgenbench2026metrics}, matched points must also have normals within $20^\circ$, which prevents matches between different faces.
\ding{185}~\emph{Edge Chamfer Distance (ECD)}~\cite{chen2020bspnet} samples 50{,}000 points on edges whose adjacent faces have normals at least $30^\circ$ apart, with distances normalized by $L$; we report it multiplied by $10^3$.
\ding{186}~\emph{DrawScore (DS)} uses the tolerance $\sigma = 0.005L$, equal to the threshold $\tau$ of surface F-score, so that the two saturating penalties are compared at the same tolerance.

\subsection{Implementation Details}
\label{sec:experiments:impl}

We register the candidate to the ground truth with the rigid registration of CADGenBench~\cite{cadgenbench2026metrics}, using one transform for all three views.
Registration preserves scale, except in RQ3, where we follow Ortho2CAD~\cite{ortho2cad} and normalize scale for both rewards and evaluation, since the model being trained often gets the scale wrong and would otherwise receive zero reward for every candidate in a training group.
We then project both models with \texttt{Open CASCADE} from a common camera at the center of the ground truth's bounding box.
To absorb small residual rotations, each view is projected along eight directions around its principal direction, each shifting the projection by less than $\sigma$.
Recall and precision are computed on each projection after merging overlapping curves of the same visibility, and averaged over the eight directions to give $R_v$ and $P_v$.
Discretization, quadrature, and tessellation settings are given in the supplementary material.
\providecolor{answerbg}{HTML}{F5F6FA}
\providecolor{answerframe}{HTML}{A9BCDB}
\providecommand{\methodname}{DrawScore}
\ProvideTColorBox{answerboxenv}{}{
  enhanced,
  colback=answerbg, colframe=answerframe,
  boxrule=0.6pt, arc=2mm,
  left=6pt, right=6pt, top=4pt, bottom=4pt,
  before skip=6pt, after skip=6pt,
}
\providecommand{\answerbox}[2]{%
  \begin{answerboxenv}
    \ding{46}~\textbf{Answer to RQ#1:} \textit{#2}
  \end{answerboxenv}}

\providecommand{\RQOneNumCases}{35}
\providecommand{\RQOneNumComparisons}{53}
\providecommand{\RQOneHoleRadii}{\{1, 2, 4, 8, 12, 16\}}
\providecommand{\RQOneHoleSlopeIoU}{3.00}
\providecommand{\RQOneHoleSlopeCD}{2.85}
\providecommand{\RQOneHoleSlopeF}{2.18}
\providecommand{\RQOneHoleSlopeECD}{0.85}
\providecommand{\RQOneHoleSlopeDS}{0.96}
\providecommand{\RQOneHoleSlopeFNoNormal}{2.18}
\providecommand{\RQOneThreadDV}{0.97}
\providecommand{\RQOneThreadDL}{83.88}
\providecommand{\RQOneThreadIoUDrop}{1.06}
\providecommand{\RQOneThreadDSDrop}{66.10}
\providecommand{\RQOneBlindDSDrop}{2.19}
\providecommand{\RQOneSaturationDisp}{20}

\section{Experimental Results}
\label{sec:experiments_results}

\subsection{RQ1: Does the integration dimension analysis hold?}
\label{sec:results:rq1}

\begin{figure}[t]
  \centering
  \includegraphics[width=\linewidth]{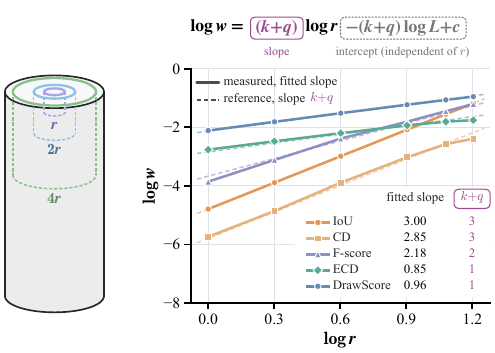}
  \caption{Weight of a local error versus its size on the blind-hole sweep of RQ1. Left: the removed blind hole of radius $r \in \RQOneHoleRadii$\,mm and depth $2r$. Right: $w$ is the metric value for CD and ECD and $1-\text{score}$ for the other metrics.}
  \label{fig:rq1_prediction}
  \vspace{-10pt}
\end{figure}

\begin{figure}[t]
  \centering
  \includegraphics[width=\linewidth]{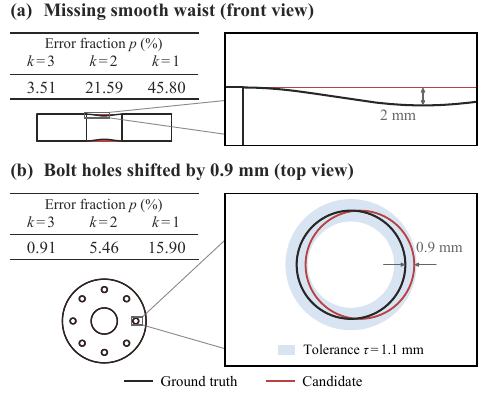}
  \caption{Representative controlled errors of RQ1: \textbf{(a)} tests the choice of curves (Finding~1) and \textbf{(b)} the saturating penalty (Finding~2). The box on the left marks the error and is magnified on the right. $p$ is the error fraction of \cref{eq:decomposition} on the solid ($k=3$), the surface ($k=2$), and the projected curves ($k=1$).}
  \label{fig:rq1_cases}
  \vspace{-10pt}
\end{figure}

\textbf{Setting.}
We test three conclusions of \cref{sec:analysis} on controlled samples: the decay order $k+q$ of \cref{eq:scaling} (Finding~2), the choice of curves in the integration domain (Finding~1), and the saturating penalty (Finding~2).
For the decay order, we remove a blind hole
(\cref{fig:rq1_prediction}, left).
The other two conclusions are each illustrated by a representative error that isolates one mechanism (\cref{fig:rq1_cases}); all \RQOneNumComparisons{} comparisons on \RQOneNumCases{} ground-truth models are listed in the supplementary material.

\textbf{Results.}
\emph{The fitted decay orders match the integration dimension analysis} (\cref{fig:rq1_prediction}).
Taking the logarithm of both sides of \cref{eq:scaling}, we test the slope against $\log r$ (\cref{fig:rq1_prediction}, right).
For all five metrics, the fitted slope differs from $k+q$ by less than 0.2. This confirms that the weight of a local error decays at the order set by the dimension of the integration domain and the penalty function.

\emph{The choice of curves separates DrawScore from ECD}.
A smooth waist creates no sharp edge, so ECD has no curve at the error and stays at 0.006, whereas the contour lines of the waist remain in the drawing (\cref{fig:rq1_cases}a) and lower DrawScore to 79.55\%.

\emph{A saturating penalty keeps deviations below the tolerance from vanishing}.
Shifting the bolt holes by 0.9\,mm keeps every point within $\tau$ (\cref{fig:rq1_cases}b), so surface F-score stays at 100\%, whereas the Gaussian penalty of DrawScore still lowers the score to 95.94\%.

\answerbox{1}{Each metric decays at the order set by its integration domain and penalty; \methodname{} additionally covers smooth outlines and keeps deviations below the tolerance visible.}

\subsection{RQ2: Do DrawScore and existing metrics capture different aspects of geometric error?}
\label{sec:results:rq2}

\begin{figure}[t]
  \centering
  \includegraphics[width=\linewidth]{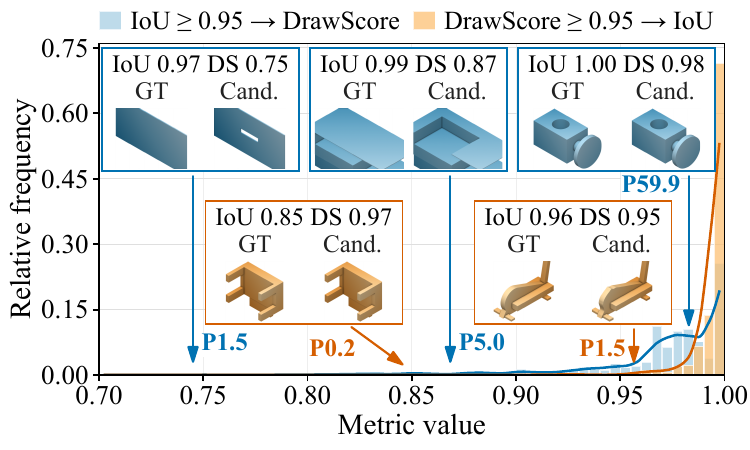}
  \caption{Bidirectional comparison of IoU and DrawScore on agent outputs (RQ2). Blue shows DrawScore for outputs with IoU $\geq 0.95$, and orange shows IoU for outputs with DrawScore $\geq 0.95$. Insets compare the GT and candidate at the marked percentiles within each subset. The horizontal axis is clipped at 0.70.}
  \label{fig:rq2_bidirectional}
  \vspace{-10pt}
\end{figure}

\textbf{Setting.}
We score all 1{,}188 valid outputs from existing CAD generation agents with the five metrics 
per output. We then compare IoU ($k=3$) and DrawScore ($k=1$)
(\cref{fig:rq2_bidirectional}). On the outputs that one metric rates as nearly perfect, that is, with IoU $\geq 0.95$ or DrawScore $\geq 0.95$, we plot the histogram of the other metric. For each histogram, we inspect selected outputs against their ground truths and mark their percentile ranks within the corresponding subset, \eg, P1.5 for the 1.5th percentile.

\textbf{Results.}
\emph{On outputs that IoU rates as nearly perfect, DrawScore retains a long lower tail.}
Among the 1{,}007 outputs with IoU $\geq 0.95$, DrawScore ranges from 0.4218 to 1.0000, and 190 of them (18.9\%) score below 0.95 (\cref{fig:rq2_bidirectional}, blue). The outputs in this tail contain local errors. 

\emph{Agent outputs tend to get the volume right before the details.}
Among the 827 outputs with DrawScore $\geq 0.95$, IoU ranges from 0.3984 to 1.0000; only 10 outputs (1.2\%) score below 0.95, while 817 (98.8\%) achieve IoU $\geq 0.95$ (\cref{fig:rq2_bidirectional}, orange).
An output that matches the drawings thus almost always matches the volume, whereas the first direction shows that the converse does not hold. The agents we evaluate therefore tend to produce outputs whose volume is nearly complete but whose remaining errors lie on the curves of the drawings, which DrawScore detects and IoU does not.

\answerbox{2}{IoU and \methodname{} capture different aspects of geometric error: \methodname{} exposes errors too small in volume for IoU, whereas outputs it rates as nearly perfect are also nearly perfect in IoU.}

\subsection{RQ3: Can \methodname{} serve as a training reward that improves generation quality?}
\label{sec:results:rq3}

\textbf{Setting.}
We perform supervised fine-tuning (SFT) on 106,220 DeepCAD samples and reinforcement learning on 6,407 Fusion360~\cite{willis2021fusion} samples, and evaluate on 153 held-out Fusion360 models.
Starting from the SFT checkpoint, we train Qwen3.5-4B with Dr.GRPO~\cite{liu2025understanding} for 400 steps; all runs share the training data and hyperparameters and differ only in the reward.
Besides \methodname{}, we select one existing metric per integration dimension as a reward.

To separate gains in validity from gains in geometry, we also run a two-stage training.
The validity reward checks only whether an output is valid: it is 1 if the generated CadQuery code executes and produces a valid solid, and 0 otherwise.
We split the reinforcement learning data into two disjoint subsets. We train with the validity reward for 200 steps on 3,203 samples, then continue with \methodname{} for 200 steps on the remaining 3,204 samples.
Training and reward details are given in the supplementary material.

\providecommand{\good}[1]{$^{\color{green!50!black}\boldsymbol{#1}}$}
\providecommand{\bad}[1]{$^{\color{red!70!black}\boldsymbol{#1}}$}
\begin{table}[t]
\centering
\caption{Reinforcement learning with different rewards (RQ3).
Superscripts give the direction of the change from SFT, in green for an improvement and in red for a degradation.
Shaded cells mark the metric that each run uses as its reward.}
\label{tab:rq3_rl}
\small
\setlength{\tabcolsep}{4pt}
\begin{adjustbox}{max width=\columnwidth}
\begin{tabular}{lccccc}
\toprule
Reward & Valid (\%)$\uparrow$ & IoU$\uparrow$ & F-score$\uparrow$ & ECD$\downarrow$ & DrawScore$\uparrow$ \\
\midrule
\textcolor{gray}{SFT} & \textcolor{gray}{75.2} & \textcolor{gray}{0.338} & \textcolor{gray}{0.307} & \textcolor{gray}{81.39} & \textcolor{gray}{0.242} \\
\midrule
IoU & 93.5\good{+} & \cellcolor{gray!20}0.484\good{+} & 0.407\good{+} & 54.29\good{-} & 0.263\good{+} \\
F1  & 85.0\good{+} & 0.439\good{+} & \cellcolor{gray!20}0.362\good{+} & 48.32\good{-} & 0.250\good{+} \\
ECD & 84.3\good{+} & 0.376\good{+} & 0.313\good{+} & \cellcolor{gray!20}71.60\good{-} & 0.220\bad{-} \\
DS  & 89.5\good{+} & 0.445\good{+} & 0.369\good{+} & 53.25\good{-} & \cellcolor{gray!20}0.272\good{+} \\
\midrule
Validity        & \cellcolor{gray!20}86.9 & 0.358 & 0.191 & 74.26 & 0.199 \\
Validity$\to$DS & 90.2\good{+} & 0.446\good{+} & 0.369\good{+} & 51.62\good{-} & \cellcolor{gray!20}0.244\good{+} \\
\bottomrule
\end{tabular}
\end{adjustbox}
\vspace{-10pt}
\end{table}

\textbf{Results.}
\cref{tab:rq3_rl} reports the results of each reward and the direction of its change from SFT.

\emph{Every geometric reward improves validity and its own metric, but rewards differ in how hard they are to learn.}
Compared with SFT, all four rewards increase the share of valid outputs and improve the metric they optimize.
The difficulty of learning differs across rewards: for example, training directly with the surface F-score reward yields a lower F-score than training with the IoU reward, indicating that surface F-score is harder to learn as a reward than IoU.

\emph{Training with \methodname{} improves overall geometric quality.}
With the \methodname{} reward, other three metrics also improve over SFT.
In contrast, the ECD reward increases validity but lowers \methodname{} below SFT.
Although both are curve metrics, ECD covers only 3D sharp edges and excludes projected contour lines and hidden lines, so the ECD reward gives no signal on these lines.

\emph{The gains do not come from validity alone.}
Training with the validity reward alone increases validity but lowers surface F-score and \methodname{} below SFT.
The model thus tends to produce outputs that execute but are geometrically coarser, and higher validity by itself does not improve geometry.
Continuing training with \methodname{} from this checkpoint improves all geometric metrics over the validity checkpoint.

\answerbox{3}{\methodname{} serves directly as a reinforcement learning reward: it improves validity and every geometric metric, beyond what validity alone brings.}

\subsection{RQ4: Can \methodname{} improve generation quality at inference time?}
\label{sec:results:rq4}

\textbf{Setting.}
The Ortho2CAD samples used in RQ2 are relatively simple: the mean IoU of every configuration lies between 0.926 and 0.967, and multiple outputs of the same model differ little, which makes it hard to test selection.
We therefore use the harder BenchCAD~\cite{benchcad2026public} and perform best-of-N selection on its 20 valid Public T1 cases.
Qwen3.8-max with the codex framework generates 8 candidates for each case, and \methodname{} compares each candidate with the input drawing and selects the one with the highest score.
We simulate the input drawing with vector projections of the ground truth in the three orthographic views.
As the baseline, we use the mean over valid candidates, i.e., single-generation performance averaged over sampling; as the upper bound (Oracle), we select the best candidate under each metric using the ground truth.

\begin{table}[t]
\centering
\caption{Best-of-N selection with \methodname{} (RQ4).}
\label{tab:rq4_tts}
\small
\setlength{\tabcolsep}{2pt}
\begin{adjustbox}{max width=\columnwidth}
\begin{tabular}{lccccc}
\toprule
Strategy & IoU$\uparrow$ & CD$\downarrow$ & F-score$\uparrow$ & ECD$\downarrow$ & DrawScore$\uparrow$ \\
\midrule
Mean of valid candidates & 0.524 & 16.19 & 0.529 & 31.79 & 0.546 \\
DrawScore selection      & 0.534 & 14.61 & 0.555 & 33.47 & 0.591 \\
\midrule
Oracle & 0.719 & 7.03 & 0.732 & 17.61 & 0.716 \\
\bottomrule
\end{tabular}
\end{adjustbox}
\vspace{-10pt}
\end{table}

\textbf{Results.}
\cref{tab:rq4_tts} reports the results of the three strategies.

\emph{\methodname{} serves as a proxy for other metrics.}
Without 3D ground truth at inference time, other metrics cannot be used for selection directly; using them would require training additional reward models to predict them.
Selecting with \methodname{} outperforms the baseline on most metrics: on IoU, CD, and surface F-score, its gains over the baseline reach 4.7\%, 17.2\%, and 13.0\% of the Oracle's gains, respectively.
By comparing candidates only with the input drawing, \methodname{} thus serves as a proxy for these metrics and removes the need to train reward models for them.

\answerbox{4}{Given a vector input drawing, best-of-N selection with \methodname{} outperforms single generation on most metrics, without 3D ground truth or an additional reward model.}

\section{Related Work}
\label{sec:related}

\paragraph{Geometric similarity metrics.}
3D reconstruction is commonly evaluated with volumetric IoU~\cite{mescheder2019occupancy}, Chamfer Distance and Earth Mover's Distance~\cite{fan2017pointset}, surface F-score~\cite{tatarchenko2019singleview}, and normal consistency~\cite{mescheder2019occupancy}.
SAUCD~\cite{luan2024saucd} compares meshes in the spectral domain with frequency-band weights learned from human ratings, and Light Field Descriptor~\cite{chen2003lightfield} compares contour descriptors from multiple projected views.
In CAD generation, SECAD-Net~\cite{li2023secad} and SfmCAD~\cite{li2024sfmcad} use Edge Chamfer Distance, which samples only sharp edges, and IterCAD~\cite{hu2026itercad} aggregates success rates over multiple Chamfer Distance thresholds.
Our analysis in \Cref{sec:analysis} unifies these metrics.
Building on it, DrawScore integrates along projected curves in engineering drawings, so the weight of a local error decays only linearly with its size.

\paragraph{CAD-specific evaluation.}
Other criteria evaluate properties beyond shape similarity.
DeepCAD~\cite{wu2021deepcad}, TransCAD~\cite{dupont2024transcad}, and CadVLM~\cite{wu2024cadvlm} compare generated modeling commands, parameters, or sketch entities with the ground-truth program.
CAD-MLLM~\cite{xu2024cadmllm} checks geometric validity such as mesh connectivity and self-intersections, CADTests~\cite{mallis2026cadtests} verifies geometric and topological requirements with sample-specific executable tests, and CADGenBench~\cite{cadgenbench2026metrics} combines shape similarity with topology and mating-interface matches.
These criteria complement geometric similarity metrics.
\section{Discussion}
\label{sec:discussion}

\subsection{Sensitivity to the Tolerance}
\label{sec:discussion:tolerance}

\begin{figure}[t]
  \centering
  \includegraphics[width=\linewidth]{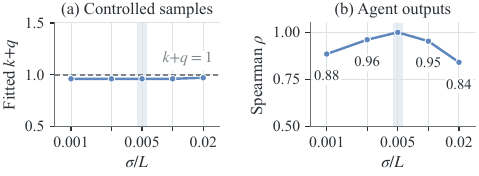}
  \caption{Sensitivity of DrawScore to the positional tolerance $\sigma$; the gray band marks the default $\sigma = 0.005L$. \textbf{(a)} Fitted decay order $k{+}q$ on the blind-hole sweep of RQ1 (\cref{fig:rq1_prediction}); the dashed line is $k{+}q=1$. \textbf{(b)} Spearman correlation between DrawScore at each $\sigma$ and at the default over the 1{,}188 agent outputs of RQ2.}
  \label{fig:disc_tolerance}
  \vspace{-10pt}
\end{figure}

DrawScore has one hyperparameter, the positional tolerance $\sigma$. We vary it twentyfold, from $0.001L$ to $0.02L$, together with the cone of angular averaging that it determines (\cref{sec:experiments:impl}), and recompute the decay order on the blind-hole sweep of RQ1 and DrawScore on the 1{,}188 agent outputs of RQ2 at each tolerance (\cref{fig:disc_tolerance}).

On the controlled samples, the fitted decay order of DrawScore stays between 0.96 and 0.97 at every tolerance, close to the order $k+q = 1$ in the integration dimension analysis. The weight of a local error thus decays linearly with its size, as determined by the integration domain and the saturating penalty rather than by the value of $\sigma$.

On the agent outputs, DrawScore at every tolerance correlates strongly with DrawScore at the default $\sigma$ (Spearman $\rho \geq 0.84$), so the relative ordering of outputs is largely preserved.
DrawScore is therefore robust to the choice of $\sigma$ within this twentyfold range.

\subsection{Extending \methodname{} with Section Views}
\label{sec:discussion:section}

For a few parts, the three orthographic views do not determine the geometry: internal structures such as stepped bores and internal cavities appear in them only as overlapping hidden lines.
Engineering drawings of such parts include additional section views, and \methodname{} can be extended in the same way.

To obtain a section view, we cut the ground truth $G$ and the candidate $C$ with a plane $\Pi$ and project them along its normal.
The section view is scored as in \cref{sec:method:pr,sec:method:aggregation} and included as another view in the minimum, so structures hidden in the three views are compared as visible lines.

Following prior work on automatic cutaways~\cite{li2007cutaway,feiner1992cutaways}, $\Pi$ can be chosen among planes parallel to the planar faces of the ground truth as the one that turns the most hidden lines into visible lines, and several such planes add several views.
We leave the evaluation of this extension to future work.

\subsection{Limitations}
\label{sec:discussion:limitations}

The test-time selection of RQ4 assumes that the input drawing is available in vector form; drawings given as raster images would first have to be vectorized. DrawScore also takes longer to compute than other metrics, since it runs hidden-line removal on each model along multiple viewing directions.

\section{Conclusion}
\label{sec:conclusion}
We showed that the weight of a local error in a geometric similarity metric is governed by the dimension of its integration domain and its penalty function. Based on this analysis, DrawScore compares the visible and hidden lines of three orthographic views with a saturating penalty. It exposes errors that existing metrics miss, improves generation as a reinforcement learning reward, and selects candidates at test time without 3D ground truth. Future work will incorporate automatically selected section views to cover internal structures.


{\small
\bibliographystyle{ieeenat_fullname}
\bibliography{11_references}
}


\end{document}